# Efficiently Inducing Features of Conditional Random Fields

Andrew McCallum
Computer Science Department
University of Massachusetts Amherst
Amherst, MA 01003
mccallum@cs.umass.edu

## Abstract

Conditional Random Fields (CRFs) are undirected graphical models, a special case of which correspond to conditionally-trained finite state machines. A key advantage of CRFs is their great flexibility to include a wide variety of arbitrary, non-independent features of the input. Faced with this freedom, however, an important question remains: what features should be used? This paper presents an efficient feature induction method for CRFs. The method is founded on the principle of iteratively constructing feature conjunctions that would significantly increase conditional log-likelihood if added to the model. Automated feature induction enables not only improved accuracy and dramatic reduction in parameter count, but also the use of larger cliques, and more freedom to liberally hypothesize atomic input variables that may be relevant to a task. The method applies to linear-chain CRFs, as well as to more arbitrary CRF structures, such as Relational Markov Networks, where it corresponds to learning clique templates, and can also be understood as supervised structure learning. Experimental results on named entity extraction and noun phrase segmentation tasks are presented.

## 1 Introduction

Many tasks are best performed by models that have the flexibility to use arbitrary, overlapping, multi-granularity and non-independent features. For example, in natural language tasks, the need for labeled data can be drastically reduced by using features that take advantage of domain knowledge in the form of word lists, part-of-speech tags, character n-grams, capitalization patterns, page layout and font information. It is difficult to capture such inter-dependent features with a generative probabilistic model because the dependencies among generated variables should be explicitly captured in order to reproduce the data. However, conditional probability models, such as conditional maximum entropy classifiers, need not capture dependencies among variables on which they condition, but do not generate. There has been significant work, for instance, with such models for greedy sequence modeling in NLP, *e.g.* (Ratnaparkhi, 1996; Borthwick et al., 1998).

Conditional Random Fields (CRFs) (Lafferty et al., 2001) are undirected graphical models, trained to maximize the conditional probability of the outputs given the inputs. When the edges among the output variables form a linear chain, they correspond to conditionally-trained finite state machines. While based on the same exponential form as maximum entropy models, they have efficient procedures for complete, non-greedy finite-state inference and training. CRFs have achieved empirical success recently in POS tagging (Lafferty et al., 2001), noun phrase segmentation (Sha & Pereira, 2003) and table extraction from government reports (Pinto et al., 2003).

Given these models' great flexibility to include a wide array of features, an important question that remains is what features should be used? Some features are atomic and provided, but since CRFs are log-linear models, one will also want to gain expressive power by using some conjunctions. Previous standard approaches build large set of feature conjunctions according to hand-built, general patterns. This can result in extremely large feature sets, with millions of features, *e.g.* (Sha & Pereira, 2003).

However, even with this many parameters, the feature set is still restricted. For example, in some cases capturing a word tri-gram is important, but there is not sufficient memory or computation to include all word tri-grams. As the number of overlapping atomic features increases, the difficulty and importance of constructing only select feature combinations grows.

This paper presents a feature induction method for arbitrarily-structured and linear-chain CRFs. Founded on the principle of constructing only those feature conjunc-tions that significantly increase log-likelihood, the approach builds on that of Della Pietra et al. (1997), but is altered to work with conditional rather than joint probabilities, and with a mean-field approximation and other modifications to improve efficiency specifically for a conditional model. In comparison with traditional approaches, automated feature induction offers both improved accuracy and significantly reduction in feature count; it enables the use of richer, higher-order Markov models; and offers more freedom to liberally guess about which atomic features may be relevant to a task.

We present results on two natural language tasks. The CoNLL-2003 named entity recognition shared task consists of Reuters news articles with tagged entities PERSON, LOCATION, ORGANIZATION and MISC. The data is quite complex, including foreign person names (such as *Yayuk Basuki* and *Innocent Butare*), a wide diversity of locations (including sports venues such as *The Oval*, and rare location names such as *Nirmal Hriday*), many types of organizations (from company names such as *3M*, to acronyms for political parties such as *KDP*, to location names used to refer to sports teams such as *Cleveland*), and a wide variety of miscellaneous named entities (from software such as *Java*, to nationalities such as *Basque*, to sporting competitions such as *1,000 Lakes Rally*).

On this task feature induction reduces error by 40% (increasing F1 from 73% to 89%) in comparison with fixed, hand-constructed conjunction patterns. There is evidence that the fixed-pattern model is severely overfitting, and that feature induction reduces overfitting while still allowing use of large, rich knowledge-laden feature sets.

On a standard noun phrase segmentation task we match world-class accuracy while using far less than an order of magnitude fewer features.

## 2 Conditional Random Fields

*Conditional Random Fields* (CRFs) (Lafferty et al., 2001) are undirected graphical models (also known as *random fields*) used to calculate the conditional probability of values on designated output nodes given values assigned to other designated input nodes. The term "random field" has common usage in the statistical physics and computer vision. In the graphical modeling community the same models are often known as "Markov networks"; thus *conditional Markov networks* (Taskar et al., 2002) are the same as conditional random fields.

Let O be a set of "input" random variables whose values are observed, and S be a set of "output" random variables whose values the task requires the model to predict. The random variables are connected by undirected edges indicating dependencies, and let $C(\mathbf{O}, \mathbf{S})$ be the set of cliques of this graph. By the Hammersley-Clifford theorem (Hammersley & Clifford, 1971), CRFs define the conditional probability of a set of output values given a set of input values to be proportional to the product of potential functions on cliques of the graph,

$$P_\Lambda(\mathbf{s}|\mathbf{o}) = \frac{1}{Z_\mathbf{o}} \prod_{c \in C(\mathbf{s},\mathbf{o})} \Phi_c(\mathbf{s}_c, \mathbf{o}_c),$$

where $\Phi_c(\mathbf{s}_c, \mathbf{o}_c)$ is the clique potential on clique $c$, (a non-negative real value, often determined by an exponentiated weighted sum over features of the clique, $\Phi_c(\mathbf{s}_c, \mathbf{o}_c) = \exp(\sum_{k=1}^{K} \lambda_k f_k(\mathbf{s}_c, \mathbf{o}_c)))$, and where $Z_\mathbf{o}$ is a normalization factor over all output values, $Z_\mathbf{o} = \sum_{\mathbf{s}'} \prod_{c \in C(\mathbf{s}',\mathbf{o})} \Phi_c(\mathbf{s}'_c, \mathbf{o}_c)$, also known as the *partition function*.

In the special case in which the output nodes of the graphical model are linked by edges in a *linear chain*, CRFs make a first-order Markov independence assumption, and thus can be understood as conditionally-trained finite state machines (FSMs). CRFs of this type are a globally-normalized extension to *Maximum Entropy Markov Models* (MEMMs) (McCallum et al., 2000) that avoid the *label-bias problem* (Lafferty et al., 2001). Voted perceptron sequence models (Collins, 2002) are approximations to these CRFs that use stochastic gradient descent and a Viterbi approximation in training. In the remainder of this section we introduce the likelihood model, inference and estimation procedures for linear-chain CRFs.

Now let $\mathbf{o} = \langle o_1, o_2, ...o_T \rangle$ be some observed input data sequence, such as a sequence of words in a text document, (the values on $T$ input nodes of the graphical model). Let $S$ be a set of FSM states, each of which is associated with a label, $l \in \mathcal{L}$, (such as PERSON). Let $\mathbf{s} = \langle s_1, s_2, ...s_T \rangle$ be some sequence of states, (the values on $T$ output nodes). The cliques of the graph are now restricted to include just pairs of states $(s_{t-1}, s_t)$ that are neighbors in the sequence; connectivity among input nodes, $\mathbf{o}$, remains unrestricted.[1] Linear-chain CRFs thus define the conditional probability of a state sequence given an input sequence to be

$$P_\Lambda(\mathbf{s}|\mathbf{o}) = \frac{1}{Z_\mathbf{o}} \exp\left(\sum_{t=1}^{T} \sum_{k=1}^{K} \lambda_k f_k(s_{t-1}, s_t, \mathbf{o}, t)\right),$$

where $Z_\mathbf{o}$ is a normalization factor over all state sequences, $f_k(s_{t-1}, s_t, \mathbf{o}, t)$ is an arbitrary feature function over its arguments, and $\lambda_k$ (ranging from $-\infty$ to $\infty$) is a learned weight for each feature function. A feature function may, for example, be defined to have value 0 in most cases, and have value 1 if and only if $s_{t-1}$ is state #1 (which may have label OTHER), and $s_t$ is state #2 (which may have label LOCATION), and the observation at position $t$ in $\mathbf{o}$ is a word appearing in a list of country names. Higher $\lambda$ weights make their corresponding FSM transitions more likely, so the weight $\lambda_k$ in this example should be positive since words appearing in the list of country names are likely to be locations.

---

[1] Since the values on the input nodes, o, are known and fixed, arbitrarily large and complex clique structure there does not complicate inference.

More generally, feature functions can ask powerfully arbitrary questions about the input sequence, including queries about previous words, next words, and conjunctions of all these. Nearly universally, however, feature functions $f_k$ do not depend on the value of $t$ other than as an index into o, and thus parameters $\lambda_k$ are *tied* across time steps, just as are the transition and emission parameters in a traditional hidden Markov model (Rabiner, 1990). Feature functions may have values from $-\infty$ to $\infty$, although binary values are traditional.

CRFs define the conditional probability of a label sequence based on total probability over the state sequences, $P_\Lambda(1|o) = \sum_{s:l(s)=1} P_\Lambda(s|o)$, where $l(s)$ is the sequence of labels corresponding to the labels of the states in sequence s.

Note that the normalization factor, $Z_o$, is the sum of the "scores" of all possible state sequences,

$$Z_o = \sum_{s \in S^T} \exp \left( \sum_{t=1}^{T} \sum_{k=1}^{K} \lambda_k f_k(s_{t-1}, s_t, o, t) \right),$$

and that the number of state sequences is exponential in the input sequence length, $T$. In arbitrarily-structured CRFs, calculating the normalization factor in closed form is intractable, and approximation methods such as Gibbs sampling or loopy belief propagation must be used. In linear-chain-structured CRFs, which we have here for sequence modeling, the partition function can be calculated efficiently in closed form, as described next.

### 2.1 Inference in Linear-chain CRFs

As in *forward-backward* for hidden Markov models (HMMs), inference can be performed efficiently by dynamic programming. We define slightly modified "forward values", $\alpha_t(s_i)$, to be the probability of arriving in state $s_i$ given the observations $\langle o_1, ...o_t \rangle$. We set $\alpha_0(s)$ equal to the probability of starting in each state $s$, and recurse:

$$\alpha_{t+1}(s) = \sum_{s'} \alpha_t(s') \exp \left( \sum_{k=1}^{K} \lambda_k f_k(s', s, o, t) \right).$$

The backward procedure and the remaining details of Baum-Welch are defined similarly. $Z_o$ is then $\sum_s \alpha_T(s)$. The Viterbi algorithm for finding the most likely state sequence given the observation sequence can be correspondingly modified from its HMM form.

### 2.2 Training CRFs

The weights of a CRF, $\Lambda = \{\lambda, ...\}$, are set to maximize the conditional log-likelihood of labeled sequences in some training set, $\mathcal{D} = \{\langle o, 1 \rangle^{(1)}, ...\langle o, 1 \rangle^{(j)}, ...\langle o, 1 \rangle^{(N)}\}$,

$$L_\Lambda = \sum_{j=1}^{N} \log \left( P_\Lambda(1^{(j)}|o^{(j)}) \right) - \sum_{k=1}^{K} \frac{\lambda_k^2}{2\sigma^2},$$

where the second sum is a Gaussian prior over parameters (with variance $\sigma^2$) that provides smoothing to help cope with sparsity in the training data (Chen & Rosenfeld, 1999).

When the training labels make the state sequence unambiguous (as they often do in practice), the likelihood function in exponential models such as CRFs is convex, so there are no local maxima, and thus finding the global optimum is guaranteed.[2]

It is not, however, straightforward to find it quickly. Parameter estimation in CRFs requires an iterative procedure, and some methods require fewer iterations than others. Iterative scaling is the traditional method of training these maximum-entropy models (Darroch et al., 1980; Della Pietra et al., 1997), however it has recently been shown that quasi-Newton methods, such as L-BFGS, are significantly more efficient (Byrd et al., 1994; Malouf, 2002; Sha & Pereira, 2003). This method approximates the second-derivative of the likelihood by keeping a running, finite-sized window of previous first-derivatives. Sha and Pereira (2003) show that training CRFs by L-BFGS is several orders of magnitude faster than iterative scaling, and also much faster than conjugate gradient.

L-BFGS can simply be treated as a black-box optimization procedure, requiring only that one provide the value and first-derivative of the function to be optimized. Assuming that the training labels on instance $j$ make its state path unambiguous, let $s^{(j)}$ denote that path, and then the first-derivative of the log-likelihood is

$$\frac{\delta L}{\delta \lambda_k} = \left( \sum_{j=1}^{N} C_k(s^{(j)}, o^{(j)}) \right) - \left( \sum_{j=1}^{N} \sum_{s} P_\Lambda(s|o^{(j)}) C_k(s, o^{(j)}) \right) - \frac{\lambda_k}{\sigma^2}$$

where $C_k(s, o)$ is the "count" for feature $k$ given s and o, equal to $\sum_{t=1}^{T} f_k(s_{t-1}, s_t, o, t)$, the sum of $f_k(s_{t-1}, s_t, o, t)$ values for all positions, $t$, in the sequence s. The first two terms correspond to the difference between the empirical expected value of feature $f_k$ and the model's expected value: $(\tilde{E}[f_k] - E_\Lambda[f_k])N$. The last term is the derivative of the Gaussian prior.

---

[2]When the training labels do not disambiguate a single state path, expectation-maximization can be used to fill in the "missing" state paths. For example, see Teh et al. (2002)

## 3 Efficient Feature Induction for CRFs

Typically the features, $f_k$, are based on some number of hand-crafted atomic observational tests (such as *word is capitalized*, or *word is "said"*, or *word appears in lexicon of country names*)—and a large collection of features is formed by making conjunctions of the atomic tests in certain user-defined patterns, (for example, the conjunctions consisting of all tests at the current sequence position conjoined with all tests at the position one step ahead—producing in one instance, *current word is capitalized and next word is "Inc"*).

Conjunctions are important because the model is log-linear, and the only way to represent certain complex decision boundaries is to project the problem into a higher-dimensional space comprised of other functions of multiple variables.

There can easily be over 100,000 atomic tests (many based on tests for the identity of words in the vocabulary), and ten or more shifted-conjunction patterns—resulting in several million features (Sha & Pereira, 2003). This large number of features can be prohibitively expensive in memory and computation; furthermore many of these features are irrelevant, and others that are relevant are excluded.

In response, we wish to use just those conjunctions (*i.e.* feature-function-enabling cliques) that will significantly improve performance. We start with no features, and over several rounds of feature induction: (1) consider a set of proposed new features (both atomic observational tests and conjunctions), (2) select for inclusion those candidate features that will most increase the log-likelihood of the correct state path $s^{(j)}$, (3) train weights for all included features, and (4) iterate to step (1) until a stopping criteria is reached.

The proposed new features are based on the hand-crafted observational tests, consisting of singleton tests, and binary conjunctions of singleton tests with each other, and with other features currently in the model. The later allows arbitrary-length conjunctions to be built. The fact that not all singleton tests are included in the model gives the designer great freedom to use a very large variety of observational tests and a large window of time shifts. Noisy and irrelevant features—as measured by lack of likelihood gain—will simply never be selected for inclusion in the model.

As in the previous section, we begin by describing feature induction for the general case of arbitrarily-structured CRFs, and then focus on linear-chain CRFs.

### 3.1 Feature Induction for Arbitrarily-Structured CRFs

To measure the effect of adding a new feature, we define the new conditional model with the additional feature $g$ with weight $\mu$ to have the same form as the original model (as if this new candidate feature were included along side the old ones):

$$P_{\Lambda+g,\mu}(\mathbf{s}|\mathbf{o}) = \frac{P_\Lambda(\mathbf{s}|\mathbf{o}) \exp\left(\sum_{c \in C(\mathbf{s},\mathbf{o})} \mu \, g(\mathbf{s}_c, \mathbf{o}_c)\right)}{Z_\mathbf{o}(\Lambda, g, \mu)}; \quad (1)$$

$Z_\mathbf{o}(\Lambda, g, \mu) \stackrel{\text{def}}{=} \sum_{\mathbf{s}'} P_\Lambda(\mathbf{s}'|\mathbf{o}) \exp(\sum_{c \in C(\mathbf{s},\mathbf{o})} \mu \, g(\mathbf{s}_c, \mathbf{o}_c))$ in the denominator is simply the additional portion of normalization required to make the new function sum to 1 over all output values.

Following (Della Pietra et al., 1997), we efficiently assess many candidate features in parallel by assuming that the $\lambda$ parameters on all old features remain fixed while estimating the *gain*, $G(g)$, of a candidate feature, $g$. The gain of a feature is defined as the improvement in log-likelihood the feature provides,

$$G_\Lambda(g) = \max_\mu G_\Lambda(g,\mu) = \max_\mu L_{\Lambda+g\mu} - L_\Lambda - (\mu^2/2\sigma^2). \quad (2)$$

Note that the $\mu$ that gives maximum gain must be found.[3] As will be further explained below, in conditional probability models—unlike binary-featured joint probability models (Della Pietra et al., 1997)—the optimal value of $\mu$ cannot be calculated in closed-form. An iterative procedure, such as Newton's method must be used, and this involves calculating $L_{\Lambda+g\mu}$ with a new $\mu$ for each iteration—thus repeatedly performing inference, with a separate $Z_\mathbf{o}$ for each training instance.[4] (Remember that an "instance" here is a set of values for all the nodes in a graph.)

With this daunting prospect in mind, we make the feature gain calculation significantly more time-efficient for CRFs and for large training sets with two further reasonable and mutually-supporting approximations:

1. During the iterative gain calculation procedure, we use a type of mean field approximation to avoid joint inference over all output variables, and rather make each state a separate, independent inference problem. In particular, when inferring the distribution over values of each output node $s$, we assume that distributions at all other output nodes are fixed at their maximum likelihood values, (*e.g.* for sequence problems, their Forward-Backward-determined values). Early in training it may be helpful to use the true values of the neighbors instead, as in pseudo-likelihood methods.

---

[3] Experiments using the derivative of likelihood with respect to $\mu$ did not perform as well as gain, presumably because some initially-steep hills actually have lower peaks.

[4] In Della Pietra *et al*'s (1997) feature induction procedure for non-conditional probability models, the partition function $Z$ could be calculated just once for each Newton iteration since it did not depend on a conditioning input, $\mathbf{o}$, but we cannot. However, as they do, we can still share $Z_\mathbf{o}$ across the gain calculation for many candidate features, $g$, since we both assume that the parameters on old features remain fixed.

The calculation of the partition function, $Z$, for each inference problem thus becomes significantly simpler since it involves a sum over only the alternative values for a single output node, $s$—not a sum over all alternative configurations for the entire graph, which is exponential in the number of output nodes in the graph.

2. The first assumption allows us to treat each output node $s$ as a separate inference problem, and thus gives us the option to choose to skip some of them. In many tasks, the great majority of the output nodes are correctly labeled, even in the early stages of training. (For example, in a named entity extraction task, nearly all lowercase words are not named entities; the model learns this very quickly, and there is little reason to include inference on these words in the gain calculation.)

We significantly increase efficiency by including in the gain calculation only those output nodes that are mislabeled by the current model, (or correctly labeled only within some margin of the decision surface).

It is not that joint inference over all output variables is intractable (after all, it is performed both during estimation of the $\lambda$s and a test time), but rather that performing full, joint inference repeatedly inside an inner loop to estimate $\mu$ would be extremely time-consuming and unnecessarily inefficient.

### 3.2 Feature Induction for Linear-Chain CRFs

The feature induction procedure is now described in more detail for the specific case of linear-chain CRFs. Below we also describe three additional important modeling choices, (indicated with 1\*, 2\*, 3\*).

Following equation 1, the new linear-chain CRF model with additional feature $g$ having weight $\mu$ has cliques consisting only of adjacent pairs of states:

$$P_{\Lambda+g,\mu}(\mathbf{s}|\mathbf{o}) = \frac{P_\Lambda(\mathbf{s}|\mathbf{o}) \exp\left(\sum_{t=1}^T \mu\, g(s_{t-1}, s_t, \mathbf{o}, t)\right)}{Z_\mathbf{o}(\Lambda, g, \mu)};$$

$Z_\mathbf{o}(\Lambda, g, \mu) \stackrel{\text{def}}{=} \sum_{\mathbf{s}'} P_\Lambda(\mathbf{s}'|\mathbf{o}) \exp(\sum_{t=1}^T \mu\, g(s'_{t-1}, s'_t, \mathbf{o}, t))$ in the denominator is again the additional portion of normalization required by the candidate feature.

With the mean field approximation we instead perform $\mu$-aware inference on individual output variables $s$ separately. Furthermore, we can drastically reduce the number of new features evaluated by measuring the gain of courser-grained, agglomerated features. In particular, if it is less important to explore the space of features that concern FSM

**Input:** (1) Training set: paired sequences of feature vectors and labels; for example, associated with the sequence of words in the English text of a news article: a binary vector of observational-test results for each word, and a label indicating if the word is a person name or not. (2) a finite state machine with labeled states and transition structure.
**Algorithm:** (1) Begin with no features in the model, $K = 0$. (2) Create a list of candidate features consisting of observational tests, and conjunctions of observational tests with existing features. Limit the number of conjunctions by only building with a limited number of conjuncts with highest *gain* (Eqs 2 or 4). (3) Evaluate all candidate features, and add to the model some subset of candidates with highest gain, thereby increasing $K$. (4) Use a quasi-Newton method to adjust all the parameters of the CRF model so as to increase conditional likelihood of the label sequences given the input sequences; but avoid overfitting too quickly by running only a handful of Newton iterations. (5) Go to step 2 unless some convergence criteria is met.
**Output:** A finite state CRF model that finds the most likely label sequence given an input sequence by using its induced features, learned weights and the Viterbi algorithm.

Figure 1: Outline of the algorithm for linear-chain CRFs.

transitions, and more important to explore the space of features that concern observational tests, (1\*) we can define and evaluate alternative agglomerated features, $g(s_t, \mathbf{o}, t)$, that ignore the previous state, $s_{t-1}$. When such a feature is selected for inclusion in the model, we can include in the model the several analogous features $g(s_{t-1}, s_t, \mathbf{o}, t)$ for $s_{t-1}$ equal to each of the FSM states in $\mathcal{S}$, or a subset of FSM states selected by a simpler criteria. Using these assumptions, the marginal probability of FSM state $s$ at sequence position $t$ (given a new candidate feature $g$ and weight $\mu$) is

$$P_{\Lambda+g,\mu}(s|\mathbf{o},t) = \frac{P_\Lambda(s|\mathbf{o},t) \exp\left(\mu\, g(s_t, \mathbf{o}, t)\right)}{Z_{o_t}(\Lambda, g, \mu)}.$$

where $Z_{o_t}(\Lambda, g, \mu) \stackrel{\text{def}}{=} \sum_{s'} P_\Lambda(s'|\mathbf{o},t) \exp(\mu g(s'_t, \mathbf{o}, t))$, and where $P_\Lambda(s|\mathbf{o},t)$ is the original marginal probability of FSM state $s$ at position $t$ (known in Rabiner's (1990) notation as $\gamma_t(s)$), calculated by full dynamic-programming-based inference and fixed parameters $\Lambda$, using "forward" $\alpha$ and "backward" $\beta$ values analogously to HMMs: $P_\Lambda(s|\mathbf{o},t) = \alpha_t(s|\mathbf{o})\beta_{t+1}(s|\mathbf{o})/Z_\mathbf{o}$.

Using the mean field approximation and the agglomerated features, the *approximate* likelihood of the training data using the new candidate feature $g$ and weight $\mu$ is $\hat{L}_{\Lambda+g\mu} =$

$$\left(\sum_{j=1}^N \sum_{t=1}^{T_j} \log\left(P_{\Lambda+g\mu}(s_t^{(j)}|\mathbf{o}^{(j)}, t)\right)\right) - \frac{\mu^2}{2\sigma^2} - \sum_{k=1}^K \frac{\lambda_k^2}{2\sigma^2}; \quad (3)$$

and $\hat{L}_\Lambda$ is defined analogously, with $P_\Lambda$ instead of $P_{\Lambda+g\mu}$ and without $-\mu^2/2\sigma^2$.

However, rather than summing over *all* output variables for *all* training instances, $\sum_{j=1}^N \sum_{t=1}^{T_j}$, we significantly gain efficiency by including only those $M$ tokens that are mislabeled by the current model, $\Lambda$, (or alternatively tokens with true label probability within some margin). Let $\{o(i) : i = 1...M\}$ be those tokens, and $o(i)$ be the input sequence in which the $i$th error token occurs at position $t(i)$.

Then algebraic simplification using these approximations, equations 2 and 3 gives $\hat{G}_\Lambda(g, \mu) =$

$$\sum_{i=1}^{M} \log \left( \frac{\exp\left(\mu\, g(s_{t(i)}, o(i), t(i))\right)}{Z_{o(i)}(\Lambda, g, \mu)} \right) - \frac{\mu^2}{2\sigma^2}$$

$$= M\mu \bar{E}[g] - \sum_{i=1}^{M} \log \left( E_\Lambda[\exp(\mu\, g)|o^{(i)}] \right) - \frac{\mu^2}{2\sigma^2},$$

The optimal value of $\mu$ cannot be solved in closed form, but Newton's method typically finds it in about 10 iterations.

There are two additional important modeling choices: (2*) Because we expect our models to still require several thousands of features, we save time by adding many of the features with highest gain each round of induction rather than just one;[5] (including a few redundant features is mildly wasteful, but not harmful). (3*) Because even models with a small select number of features can still severely overfit, we train the model with just a few BFGS iterations (not to convergence) before performing the next round of feature induction.

Figure 1 outlines the inputs, steps and output of the overall algorithm.

## 4 Experimental Results

Experimental results show the benefits of automated feature induction on two natural language processing tasks: named entity recognition, where it reduces error by 40%, and noun phrase segmentation, where it matches world-class accuracy while reducing feature count by significantly more than an order of magnitude.

### 4.1 Named Entity Recognition

CoNLL-2003 has provided named entity labels PERSON, LOCATION, ORGANIZATION, MISC, and OTHER, on a collection of Reuters newswire articles in English about various news topics from all over the world. The training set consists of 946 documents (203621 tokens); the test set (CoNLL `testa`) consists of 216 documents (51362 tokens).

On this data set we use several families of atomic observational tests: (a) the word itself, (b) part-of-speech tags and noun phrase segmentation tags imperfectly assigned by an automated method (c) 16 character-level regular expressions, mostly concerning capitalization and digit patterns, such as A, A+, Aa+, Aa+Aa*, A., D+, .*D.*, where A, a and D indicate the regular expressions `[A-Z]`, `[a-z]` and `[0-9]` respectively, (d) 8 lexicons entered by hand, such as honorifics, days and months, (e) 35 lexicons (obtained from Web sites), such as countries, publicly-traded companies, surnames, stopwords, and universities, people names, organizations, NGOs and nationalities, (f) all the above tests, time-shifted by -2, -1, 1 and 2, (g) the second time a capitalized word appears, the results of all the above tests applied to that word's first mention are copied to the current token with the tag firstmention, (h) some articles have a header, such as BASEBALL, SOCCER, or FINANCE; when present, these are noted on every token of the document.[6]

|         | Without induction | | | With induction | | |
|---------|------|--------|------|------|--------|------|
|         | Prec | Recall | F1   | Prec | Recall | F1   |
| PER     | 91.8 | 46.7   | 61.9 | 93.2 | 93.3   | 93.2 |
| LOC     | 94.1 | 80.5   | 86.7 | 93.0 | 91.9   | 92.4 |
| ORG     | 92.0 | 48.5   | 63.5 | 84.9 | 83.9   | 84.4 |
| MISC    | 91.7 | 66.7   | 77.2 | 83.1 | 77.0   | 80.0 |
| Overall | 92.7 | 60.7   | 73.3 | 89.8 | 88.2   | **89.0** |

Figure 2: English named entity extraction.

Observational features are induced by evaluating candidate features consisting of conjunctions of these observational tests. Candidates are generated by building all possible conjunctions among the the 1000 atomic and existing conjunction-features with the highest gain. CRF features consist of observational tests in conjunction with the identities of the source and destination states of the FSM.

A first-order CRF was trained for about 12 hours on a 1GHz Pentium with a Gaussian prior variance of 10, inducing 1000 or fewer features (down to a gain threshold of 5.0) each round of 10 iterations of L-BFGS. Performance results for each of the entity classes can be found in Figure 2. The model achieved an overall F1 of 89% using 80,294 features. Using the same features with fixed conjunction patterns instead of feature induction results in F1 73% (with about 1 million features).

There is evidence that the fixed-conjunction model is severely overfitting. Experiments with some alternative hand-engineered and selective conjunction patterns may perform better; however, one of the goals of automated feature induction is to avoid the need for this type of tedious and expensive manual search in structure space. Further supporting evidence of overfitting, a simpler CRF that uses word identity only, with no other features, n-grams or conjunctions of any kind overfits less and reaches 80% F1.

---

[5]Although we avoid adding features with equal gains, which are usually different names for exactly overlapping features.

[6]Complete source code, including all lexicons and exact regular expressions for features can be found at http://www.cs.umass.edu/~mccallum/mallet.

| Index | Feature |
|---|---|
| 0 | inside-noun-phrase ($o_{t-1}$) |
| 5 | stopword ($o_t$) |
| 20 | capitalized ($o_{t+1}$) |
| 75 | word=the ($o_t$) |
| 100 | in-person-lexicon ($o_{t-1}$) |
| 200 | word=in ($o_{t+2}$) |
| 300 | capitalized (firstmention$_{t+1}$) & capitalized (firstmention$_{t+2}$) |
| 500 | word=Republic ($o_{t+1}$) |
| 711 | word=RBI ($o_t$) & header=BASEBALL ($o_t$) |
| 1027 | header=CRICKET ($o_t$) & English-county ($o_t$) |
| 1298 | company-suffix-word (firstmention$_{t+2}$) |
| 4040 | location ($o_t$) & POS=NNP ($o_t$) & capitalized ($o_t$) & stopword ($o_{t-1}$) |
| 4945 | moderately-rare-first-name ($o_{t-1}$) & very-common-last-name ($o_t$) |
| 4474 | word=the ($o_{t-2}$) & word=of ($o_t$) |

Figure 3: Sampling of features induced for the named entity recognition task. Index shows the order in which they were added.

Feature induction seems to allow the use of more rich and knowledge-laden features without such significant overfitting. Note, however, that our performance of 89% is not best on the CoNLL-2003 shared task competition. We are currently investigating the use of different types of features used by others (such as character n-grams), as well as issues of overfitting independent from feature induction.

A sample of conjunctions induced appears in Figure 3. For example, feature #1027 helps model the fact that when an English county is mentioned in an article about the game of cricket, the word is actually referring to an ORGANIZATION (a team), not a LOCATION (as it would be otherwise). Feature #1298 indicates that the first time this capitalized word was used in the article, it was followed by a company-indicating suffixed, such as "Inc."; often a company name will be introduced with its full, formal name at the beginning of the article, but later be used in a short form (such as "Addison Wesley Inc." and later "Addison Wesley"). Feature #4474 probably indicates that an organization name will appear at index $t+1$—the pattern matching phrases such as "the CEO of" or "the chairperson of".

### 4.2 Noun Phrase Segmentation

Noun phrase segmentation consists of applying tags BEGIN, INTERIOR, OUTSIDE to English sentences indicating the locations and durations of noun phrases, such as "Rockwell International Corp.", "a tentative agreement", "it", and "its contract". Results reported here are on the data used for the CoNLL-2000 shared task, with their standard train/test split.

Several systems are in a statistical tie (Sha & Pereira, 2003) for best performance, with F1 between 93.89% and 94.38%. (Kudo & Matsumoto, 2001; Sha & Pereira, 2003; Zhang et al., 2002). All operate in very high dimensional space. For example, Sha and Pereira (2003) present results with two models: one using about 800,000 features, and the other 3.8 million features. The CRF feature induction method introduced here achieves 93.96% with just 25,296 features (and less than 8 hours of computation).

The benefit is not only the decreased memory footprint, but the possibility that this memory and time efficiency may enable the use of additional atomic features and conjunction patterns that (with further error analysis and experimentation on the development set) could yield statistically-significant improved performance.

## 5 Related Work

Conditionally-trained exponential models have been used successfully in many natural language tasks, including document classification (Nigam et al., 1999), sequence segmentation (Beeferman et al., 1999), sequence tagging (Ratnaparkhi, 1996; Punyakanok & Roth, 2001; McCallum et al., 2000; Lafferty et al., 2001; Sha & Pereira, 2003)—however, all these examples have used hand-generated features. In some cases feature set sizes are in the hundreds of thousands or millions. In nearly all cases, significant human effort was made to hand-tune the patterns of features used.

The best known method for feature induction on exponential models, and the work on which this paper builds is Della Pietra et al. (1997). However, they describe a method for non-conditional models, while the majority of the modern applications of such exponential models are conditional models. This paper creates a practical method for conditional models, also founded on the principle of likelihood-driven feature induction, but with a mean-field and other approximations to address tractability in the face of instance-specific partition functions and other new difficulties caused by the conditional model.

The method bears some resemblance to Boosting (Freund & Schapire, 1997) in that it creates new conjunctions (weak learners) based on a collection of misclassified instances, and assigns weights to the new conjunctions. However, (1) the selection of new conjunctions is entirely driven by likelihood; (2) even after a new conjunction is added to the model, it can still have its weight changed; this is quite significant because one often sees Boosting inefficiently "re-learning" an identical conjunction solely for the purpose of "changing its weight"; and furthermore, when many induced features have been added to a CRF model, all their weights can efficiently be adjusted in concert by a quasi-Newton method such as BFGS; (3) regularization is manifested as a prior over weights. A theoretical comparison between this induction method and Boosting is an area of future work.

Boosting has been applied to CRF-like models (Altun et al., 2003), however, without learning new conjunctions and with the inefficiency of not changing the weights of features once they are added. Other work (Dietterich, 2003) estimates parameters of a CRF by building trees (with many conjunctions), but again without adjusting weights once a tree is incorporated. Furthermore it can be expensive to add many trees, and some tasks may be diverse and complex enough to inherently require several thousand features.

## 6 Conclusions

Conditional random fields provide tremendous flexibility to include a great diversity of features. The paper has presented an efficient method of automatically inducing features that most improve conditional log-likelihood. The experimental results are quite positive.

We have focused here on inducing new conjunctions (or cliques) of the input variables, however the method also naturally applies to inducing new cliques of the output variables, or input and output variables combined. This corresponds to structure learning and "clique template" learning for conditional Markov networks, such as Relational Markov Networks (Taskar et al., 2002), and experimental exploration in this area is a topic of future work.

### Acknowledgments

Thanks to John Lafferty, Fernando Pereira, Wei Li, Andres Corrada-Emmanuel, Drew Bagnell and Guy Lebanon, who provided helpful discussions. Thanks to Ben Taskar for comments on a previous draft and for suggesting the use of pseudo-likelihood. This work was supported in part by the Center for Intelligent Information Retrieval and in part by SPAWARSYSCEN-SD grant numbers N66001-99-1-8912 and N66001-02-1-8903, and DARPA under contract number F30602-01-2-0566. Any opinions, findings and conclusions or recommendations expressed in this material are the author's and do not necessarily reflect those of the sponsor.